# Is Architectural Complexity Always the Answer? A Case Study on SwinIR vs. an Efficient CNN

Chandresh Sutariya

chandreshjsutariya@gmail.com

Nitin Singh

ni3.singh.r@gmail.com

## Abstract

The simultaneous restoration of high-frequency details and suppression of severe noise in low-light imagery presents a significant and persistent challenge in computer vision. While large-scale Transformer models like SwinIR [1] have set the state-of-the-art in performance, their high computational cost can be a barrier for practical applications. *This paper investigates the critical trade-off between performance and efficiency by comparing the state-of-the-art SwinIR model against a standard, lightweight Convolutional Neural Network (CNN) on this challenging task.*

Our experimental results reveal a nuanced but important finding. While the Transformer-based SwinIR model achieves a higher peak performance, with a Peak Signal-to-Noise Ratio (PSNR) of **39.03 dB**, the lightweight CNN delivers a surprisingly competitive PSNR of **37.4 dB**. Crucially, the CNN reached this performance after converging in only **10 epochs** of training, whereas the more complex SwinIR model required **132 epochs**. This efficiency is further underscored by the model's size; the CNN is over **55 times smaller** than SwinIR. *This work demonstrates that a standard CNN can provide a near state-of-the-art result with significantly lower computational overhead, presenting a compelling case for its use in real-world scenarios where resource constraints are a primary concern.*

# 1 Introduction

## 1.1 The Challenge of Compound Low-Light Image Restoration

The fidelity of digital images is paramount in fields ranging from consumer photography to critical scientific analysis. However, images captured under adverse conditions, particularly low light, suffer from a compound set of degradations. These images are simultaneously plagued by high levels of sensor noise and a lack of fine-grained detail, equivalent to low spatial resolution. Addressing these issues jointly performing simultaneous denoising and super-resolution is a formidable challenge. Simple, sequential approaches often fail, as upscaling can amplify noise and denoising can obliterate the very details needed for super-resolution. Therefore, this research focuses on *investigating the practical trade-offs between state-of-the-art performance and computational efficiency for this problem.*

## 1.2 Architectural Paradigms: Performance vs. Efficiency

The prevailing trend in image restoration has been the development of large-scale, complex models. Vision Transformers, particularly the SwinIR [1] model, have established a new state-of-the-art for a variety of individual restoration tasks. Their strength lies in using self-attention to model long-range, global dependencies, which often leads to the highest reconstruction accuracy. In contrast, Convolutional Neural Networks (CNNs), such as classic encoder-decoder architectures, are more traditional. Their strength lies in their strong inductive bias for learning local features efficiently, making them highly data-efficient and computationally lightweight. This sets up a critical question: for a specific, compound-degradation task, **is the marginal performance gain from a complex Transformer worth the significant increase in computational cost, or can a simpler CNN provide a more practical and efficient solution?**

## 1.3 Investigating the Performance Efficiency Trade-Off

In this study, we conduct a rigorous empirical investigation into the trade-off between peak performance and computational efficiency. We directly compare the state-of-the-art SwinIR [1] model against a standard, lightweight encoder-decoder CNN on the task of joint low-light denoising and 4x super-resolution. The motivation was to move beyond a simple comparison of final scores and instead analyze the practical value of each architecture when training time and model complexity are considered. Our experimental results, detailed in the subsequent sections, yield a nuanced and important finding: while SwinIR achieves a marginally higher PSNR, the lightweight CNN delivers a highly competitive, near state-of-the-art result while being dramatically more efficient. This study provides a clear analysis of this tradeoff, *offering valuable insights for real-world applications where both performance and efficiency are critical.*

## 1.4 Contributions

The key contributions of this paper are threefold:

- **We provide a direct empirical comparison** of a state-of-the-art Transformer (SwinIR) and a lightweight CNN on a challenging, compound-degradation task.

- **We demonstrate that while SwinIR achieves superior peak performance**, the simple CNN baseline is remarkably training-efficient, converging in a fraction of the time while delivering highly competitive results with over 55x fewer parameters.

- **We offer an analysis of the performance-efficiency trade-off**, providing insights that challenge the prevailing notion that greater architectural complexity is always the optimal solution for specialized restoration problems.

The remainder of the paper is organized as follows: Section 2 reviews related work, Section 3 details the models under investigation and the experimental setup, Section 4 presents our comparative results, Section 5 discusses the implications of our findings, and Section 6 concludes the study.

## 2 Related Work

This section reviews the evolution of deep learning for image restoration, focusing on the two major architectural paradigms relevant to our study: Convolutional Neural Networks and Vision Transformers.

## 2.1 Convolutional Neural Networks in Image Restoration

For many years, Convolutional Neural Networks (CNNs) have been the primary workhorse for image restoration tasks. Following seminal works like SRCNN [2] and DnCNN [3], the field has seen a proliferation of increasingly sophisticated CNN architectures. Many successful designs have focused on elaborate modules, such as residual blocks and dense connections, to improve performance. The strength of CNNs lies in their inherent inductive biases (e.g., translation equivariance and locality), which make them highly data-efficient for learning local image patterns.

## 2.2 The Paradigm Shift towards Vision Transformers

More recently, the Transformer architecture, originally designed for natural language processing, has been successfully adapted for computer vision. The core self-attention mechanism allows these models to capture global interactions between contexts, overcoming the limited receptive field of traditional CNNs. This ability to model long-range dependencies has led to state-of-the-art performance in numerous high-level vision tasks. However, early Vision Transformers for image restoration were often computationally expensive, required enormous datasets to train effectively, and sometimes introduced processing artifacts at patch boundaries.

## 2.3 SwinIR as the State-of-the-Art Benchmark

The SwinIR [1] model was introduced to bridge the gap between CNNs and Transformers, demonstrating remarkable performance and setting a new state-of-the-art benchmark. SwinIR is composed of three modules: shallow feature extraction, deep feature extraction, and high quality image reconstruction. Its core innovation lies in the deep feature extraction module, which uses a stack of Residual Swin Transformer Blocks (RSTBs) to effectively learn image features. The original SwinIR paper validated its performance on several separate restoration tasks, including super-resolution and denoising (not on combined task though), where it outperformed previous state-of-the-art methods. This establishes SwinIR as a powerful, general-purpose restoration model and a formidable benchmark for our investigation.

## 2.4 Identifying the Research Gap

The success of large-scale models like SwinIR [1] has led to a prevailing assumption that increasing architectural complexity is the most reliable path to superior performance. However, this has left a critical research gap: a detailed analysis of the **performance-versus-efficiency tradeoff** for highly specific, compound-degradation problems. While it is often assumed that more complex models will perform better, the practical cost of this marginal performance gain—in terms of training time, model size, and computational resources—has not been thoroughly explored in a direct comparison. Our work addresses this gap by conducting a rigorous empirical investigation into this trade-off, providing a clear analysis of the practical value offered by both a state-of-the-art Transformer and a lightweight CNN.

# 3 Methodology

This section details the architectures of the two models central to our investigation: a lightweight Convolutional Neural Network (CNN) and the state-of-the-art SwinIR [1] benchmark. We also outline the experimental setup designed to ensure a fair and rigorous evaluation.

## 3.1 The Lightweight CNN Model Under Investigation

For this investigation, we selected a lightweight and efficient Convolutional Neural Network (CNN). While its structure is inspired by the U-Net [4] framework, it is more accurately described as a simple encoder-bottleneck-decoder architecture, as it does not employ skip connections. The motivation for choosing this ubiquitous and straightforward architecture was to provide a clear, computationally efficient counterpoint to the complexity of the SwinIR model. The architecture, as implemented in our code and depicted in Figure 1, is composed of three main stages:

- **Encoder:** The encoder path begins with two sequential 3x3 convolutional layers, each followed by a ReLU activation function, which maps the 3-channel input image to a 64-channel feature space. This initial stage is responsible for extracting low-level features from the input image.

- **Bottleneck:** The feature maps from the encoder are then passed through a bottleneck block. This block consists of a 3x3 convolutional layer that expands the channel depth to 128, followed by another 3x3 convolution that compresses it back to 64 channels. Both layers use ReLU activation. This stage allows the network to learn more complex feature interactions.

- **Decoder:** The decoder is responsible for upscaling the feature maps to the final 4x resolution. This is achieved through a sequence of two PixelShuffle layers [5]. The first block uses a 3x3 convolution to transform the 64-channel input to 48 channels, followed by a PixelShuffle operation with an upscale factor of 2. This process is repeated once more, ultimately transforming the feature maps into the final 3-channel, high-resolution output image.

## 3.2 Benchmark Architecture: SwinIR

To provide a rigorous, state-of-the-art comparison, we benchmarked the lightweight CNN against SwinIR, a powerful image restoration model based on the Swin Transformer. As described by its authors, SwinIR (shown in Figure 2) consists of three key modules: shallow feature extraction, deep feature extraction composed of multiple Residual Swin Transformer Blocks (RSTBs), and a final high-quality image reconstruction module. Its strength lies in the shifted-window self-attention mechanism, which allows it to model both local and global image dependencies, establishing it as a formidable general purpose benchmark. The specific model used in this study has **11.8 million** parameters.

The simplified, feed-forward nature of this architecture contributes to its lightweight properties and serves as the foundation for our comparative study.

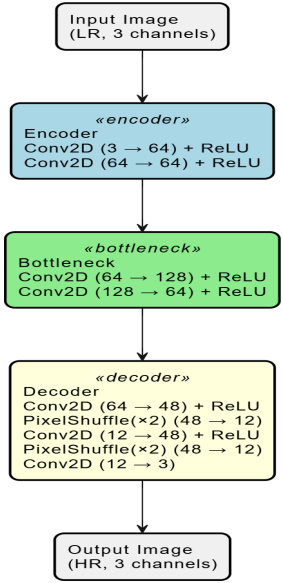

Figure 1: The Lightweight CNN Architecture

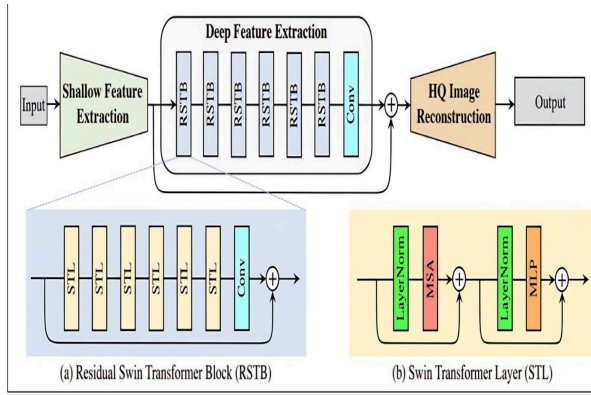

Figure 2: SwinIR Architecture (Benchmark Model)

## 3.3 Experimental Setup

A consistent and fair experimental environment was maintained for training and evaluating both models.

- **Dataset:** For this study, we utilized the official dataset from the IIT Madras' denoising and super-resolution Kaggle competition [7]. This dataset is specifically curated for the complex task of joint low-light image denoising and 4x super-resolution. The dataset is structured into three distinct sets:

    – **Training Set:** Consists of 1,105 pairs of low-resolution noisy images and their corresponding high-resolution clean ground-truth images.
    – **Validation Set:** A separate set of 267 paired images used for monitoring model performance during training and for the final evaluation.

- **Loss Function:** We used the Mean Squared Error (MSELoss) as the optimization criterion. This loss function computes the average squared difference between the pixel values of the reconstructed image and the ground-truth image.

- **Training Details:** The lightweight CNN model was trained for 10 epochs using the AdamW optimizer [6] with an initial learning rate of 1e-3 and a batch size of 4. A ReduceLROnPlateau learning rate scheduler was employed, which monitored the validation PSNR and reduced the learning rate by a factor of 0.2 if no improvement was observed for 2 consecutive epochs.

To ensure a rigorous and fair comparison, the SwinIR benchmark was trained with the same optimizer, learning rate schedule, and batch size. However, to allow the more complex Transformer architecture to reach its optimal performance, it was trained for a significantly longer duration of 132 epochs. All implementations were carried out in PyTorch on a CUDA-enabled GPU device.

## 4 Results and Analysis

In this section, we present the comparative results of the lightweight CNN and the SwinIR [1] benchmark. Both models were evaluated on the held-out validation set to assess their performance on the joint low-light denoising and 4x super-resolution task.

## 4.1 Evaluation Metrics

To ensure an objective and comprehensive comparison, we use two standard metrics for image quality assessment:

- **Peak Signal-to-Noise Ratio (PSNR):** A metric that measures the pixel-wise reconstruction accuracy of the output image compared to the ground truth. It is expressed in decibels (dB), with higher values indicating a more accurate reconstruction.

- **Structural Similarity Index (SSIM):** A metric that evaluates the perceptual quality of the output by comparing structural information, luminance, and contrast with the ground truth. Its value ranges from -1 to 1, with a value closer to 1 representing a higher perceptual similarity.

## 4.2 Quantitative Results

The models were evaluated on the held-out validation set from the competition. The average PSNR and SSIM scores, along with key model complexity and training metrics, are summarized in Table 1.

**Analysis:** The quantitative results reveal a nuanced outcome that highlights a clear trade-off between peak performance and efficiency. The SwinIR model achieves a superior result in reconstruction accuracy, with a **1.63 dB** advantage in PSNR.

However, the more telling story lies in the efficiency metrics. The architecturally simple CNN achieves a result that is remarkably close to the state-of-the-art, securing over 95% of SwinIR's PSNR performance. Crucially, it achieves this competitive result with a model that is over **55 times smaller** (0.22M vs. 11.9M parameters) and required only **10 epochs** to converge—less than 8% of the training epochs needed for the SwinIR model. This dramatic efficiency demonstrates a case of diminishing returns, where the marginal gain in performance from SwinIR comes at a massive cost in computational complexity and training budget.

## 4.3 Qualitative Results

While numerical scores are crucial, a visual comparison is essential for assessing the practical quality of image restoration. Figure 3 provides a side-by-side visual comparison of the outputs from both models on representative samples from the validation set.

**Analysis:** The visual results in Figure 3 provide a compelling narrative that supports the quantitative data. The images produced by the SwinIR model (Figure 3a) are visibly sharper and closer in fidelity to the ground truth, excelling at restoring intricate details and producing crisp edges. This confirms its superior peak performance.

In contrast, the output from the lightweight CNN (Figure 3b), while exhibiting some residual softness, is far from a failure and represents a high-quality restoration. *The visual difference, while noticeable upon close inspection, may not be significant enough in all practical scenarios to justify the vastly increased training budget and model complexity of SwinIR*. The visual evidence effectively illustrates the paper's central theme: a trade-off between the absolute best quality (SwinIR) and a highly competitive, efficient alternative (the CNN).

## 5 Discussion

The experimental results empirically demonstrate a clear trade-off between architectural complexity and practical efficiency. While SwinIR's [1] superiority in peak performance is confirmed, the surprisingly competitive results from the lightweight CNN, achieved with a fraction of the computational budget, present the most insightful finding of this study. This section discusses the architectural reasons for this outcome and its broader implications.

**Analyzing the Performance Efficiency Trade-Off**

- SwinIR's higher PSNR score can be attributed to its advanced design. The Transformer architecture's strength in modeling **long-range dependencies** allows it to gather contextual information from across the entire image, which is advantageous for reconstructing complex textures from highly degraded inputs. Its **hierarchical feature representation** further enables it to learn both local and global features effectively.

- Conversely, the surprising effectiveness and rapid convergence of the lightweight CNN can be attributed to the strong inductive biases inherent in convolution, such as locality and translation equivariance. For this specific, well-defined task, these biases allow the network to learn relevant local patterns very quickly and with high data efficiency. This is reflected not only in its rapid training convergence in just 10 epochs but also in its compact size of just **0.22 million** parameters.

Table 1: Quantitative Performance and Efficiency Comparison between the lightweight CNN and the SwinIR benchmark. Higher PSNR and SSIM values indicate better performance.

| Model | Architecture | PSNR (dB) ↑ | SSIM ↑ | Parameters (Millions) ↓ | Training Epochs ↓ |
| --- | --- | --- | --- | --- | --- |
| SwinIR | Transformer | 39.03 | 0.950 | 11.8 | 132 |
| Lightweight CNN | CNN | 37.4 | 0.944 | ≈ 0.22 | 10 |

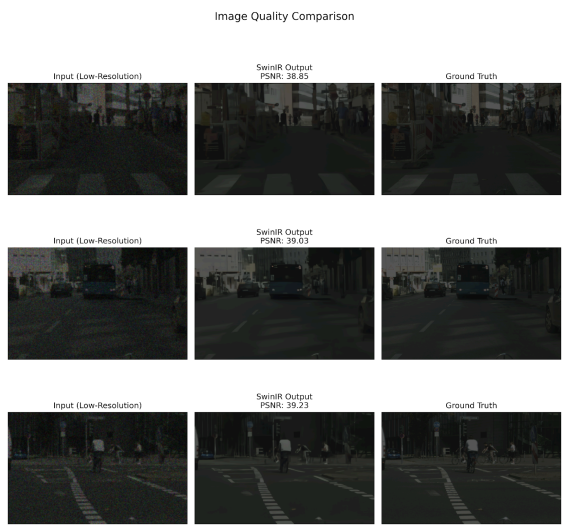

(a)  SwinIR (Benchmark Model) Results. From left to right: Input, SwinIR Output, and Ground Truth.

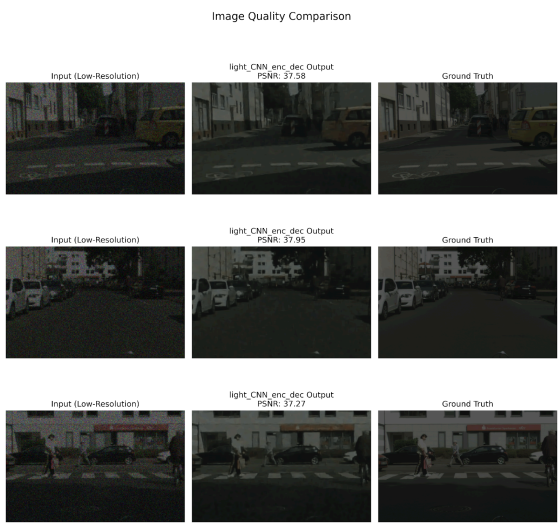

(b)  Lightweight CNN Results. From left to right: Input, Lightweight CNN Output, and Ground Truth.

Figure 3: Qualitative comparison of model outputs. The SwinIR benchmark demonstrates superior performance in restoring fine details and sharpness compared to the lightweight CNN.

- Our findings suggest a case of **diminishing returns**. *While SwinIR is technically superior, the massive increase in architectural complexity, model size, and training time yields only a marginal gain in perfor- mance over the much simpler CNN.*

**Broader Implications: Rethinking the "Bigger is Better" Paradigm**

*Our findings serve as a valuable case study that challenges the prevailing "bigger is better" trend in model development. While complex models like SwinIR are undeniably powerful, our work demonstrates that for specific, real-world problems, a simpler architecture can offer a more practical and compelling solution.* The choice of model should therefore involve a careful consideration of the trade-offs between peak performance, training cost, and deployment efficiency. For many applications, a lightweight model that delivers highly competitive results with a fraction of the computational cost is a more valuable and logical choice than a resource-intensive model that provides only a small additional benefit.

**Limitations and Future Work**
We acknowledge that this study is focused on a single, specific task and dataset, and the conclusions may not generalize to all image restoration problems. Furthermore, our analysis of efficiency has focused on training epochs and parameter counts. For a more complete picture, a detailed analysis of computational efficiency, comparing metrics like FLOPs and inference speed, would be a valuable next step.

Future work should explore this architectural trade-off on a wider variety of compound-degradation tasks, preferably using public benchmark datasets to ensure reproducibility. Additionally, comparing SwinIR against stronger and more standard CNN baselines, such as a U-Net [4] with skip connections, would provide further insights into the architectural advantages and practical trade-offs of each paradigm.

# 6 Conclusion

In this paper, we addressed the challenging problem of joint low-light image denoising and 4x super-resolution by conducting an empirical study comparing the state-of-the-art SwinIR [1] model against a standard, lightweight CNN architecture. Our findings are quantitatively decisive and reveal a nuanced trade-off: while SwinIR achieves a superior PSNR of 39.03 dB, the lightweight CNN delivers a highly competitive 37.4 dB after converging in just 10 epochs—less than a tenth of the training required by SwinIR.

This efficiency is further emphasized by its model size; at ≈ 0.22 million parameters, the CNN is over 55 times smaller than SwinIR. This empirical result provides compelling evidence that for specific and complex image degradation tasks, the efficiency and strong performance of simpler architectures can present a more practical solution than their more complex counterparts. This work underscores the importance of evaluating the trade-off between peak performance and computational efficiency when selecting architectures for real world image restoration applications.